\documentclass[lettersize,journal]{IEEEtran}
\usepackage[colorlinks,
          linkcolor=blue,
          anchorcolor=green,
          urlcolor=blue,
          citecolor=blue,
          ]{hyperref}
\usepackage{adjustbox}
\usepackage[ruled,linesnumbered]{algorithm2e}
\ifCLASSOPTIONcompsoc
\usepackage[caption=false, font=normalsize, labelfont=sf, textfont=sf]{subfig}
\else
\usepackage[caption=false, font=footnotesize]{subfig}
\fi
\usepackage{url}
\usepackage[normalem]{ulem}

\usepackage[flushleft]{threeparttable}
\usepackage{booktabs}
\usepackage{makecell}
\usepackage{multirow}
\usepackage{graphicx}

\usepackage{amsthm,amsmath,amssymb}
\newtheorem{theorem}{Masking Rule}
\usepackage{mathrsfs}
\usepackage{cleveref}
\usepackage{cite}

\def\BibTeX{{\rm B\kern-.05em{\sc i\kern-.025em b}\kern-.08em
    T\kern-.1667em\lower.7ex\hbox{E}\kern-.125emX}}
\usepackage{balance}

\newcounter{subeq}

\begin{document}
\title{An End-to-End Learning Approach for Solving Capacitated Location-Routing Problem}

\author{Chang-Hao Miao, Yun-Tian Zhang, Tong-Yu Wu, \\Fang Deng, \emph{Fellow}, \emph{IEEE} and Chen Chen, \emph{Member}, \emph{IEEE}\thanks{This work was supported by the National Key Research and Development Program of China No.2022ZD0119703; in part by the National Natural Science Foundations of China (NSFC) under Grant 62273044 and 62022015; in part by the National Natural Science Foundation of China National Science Fund for Distinguished Young Scholars under Grant 62025301; in part by the National Natural Science Foundation of China Basic Science Center Program under Grant 62088101. \emph{(Corresponding authors: Chen Chen)}}
\thanks{Chang-Hao Miao, Yun-Tian Zhang, Tong-Yu Wu, Fang Deng, and Chen Chen are with the National Key Lab of Autonomous Intelligent Unmanned Systems, Beijing Institute of Technology, Beijing 100081, China (e-mail: xiaofan@bit.edu.cn).}}

\markboth{\parbox[t]{\textwidth}{This work has been submitted to the IEEE for possible publication.\newline Copyright may be transferred without notice, after which this version may no longer be accessible.}}{How to Use the IEEEtran \LaTeX \ Templates}

\maketitle

\begin{abstract}
The capacitated location-routing problem (CLRP) is a classical problem in combinatorial optimization, which requires simultaneously making location and routing decisions. In CLRP, the complex constraints and the intricate relationships between various decisions make the problem challenging to solve. With the emergence of deep reinforcement learning (DRL), it has been extensively applied to address the vehicle routing problem and its variants, while the research related to CLRP still needs to be explored. In this paper, we propose the DRL with heterogeneous query (DRLHQ) to solve the CLRP. To the best of our knowledge, this is the first end-to-end learning framework for CLRP based on an encoder-decoder architecture, which jointly learns location and routing decisions within a unified model. In particular, we reformulate the CLRP as a markov decision process tailored to various decisions, a general modeling framework that can be easily adapted to other DRL-based methods. To better handle the interdependency across location and routing decisions, we further propose a state-conditioned heterogeneous querying attention mechanism that constructs decision-specific queries from aggregated state representations and dynamically adapts across decision stages. Experimental results on both synthetic and benchmark datasets demonstrate superior solution quality and better generalization performance of our proposed approach over representative traditional and DRL-based baselines. 
\end{abstract}

\begin{IEEEkeywords}
Location-routing problem, deep reinforcement learning, end-to-end, combinatorial optimization.
\end{IEEEkeywords}

\section{Introduction}
\IEEEPARstart{T}{he} facility location problem (FLP) and vehicle routing problem (VRP) are two critical combinatorial optimization problems (COPs) in transportation and logistics, which are traditionally addressed sequentially. However, planning the routes after facility location may lead to suboptimal solutions due to the interdependencies across various decisions \cite{salhi1989effect,salhi1999consistency}. Therefore, the capacitated location-routing problem (CLRP) \cite{cooper1972transportation} is proposed to simultaneously make location and routing decisions. The CLRP is one of the most classical topics in the community of operations research and have extensive applications such as supply-chain management \cite{galindres2023multi}, emergency management \cite{xu2024flexible}, and disaster relief \cite{zhen2014disaster}. 

Generally, the CLRP is required to jointly determine where to locate and how to route, minimizing the total cost of facility opening, vehicle routing, and vehicle operation. In the CLRP, depots and vehicles are subject to the maximum capacity constraints, and the depots are considered heterogeneous due to distinct capacities and opening costs. It should be mentioned that the CLRP belongs to non-deterministic polynomial hard (NP-hard) COPs \cite{prodhon2014survey}. Traditional methods for solving the CLRP mainly include exact and heuristic methods. Exact methods often exhibit exponential computational complexity, which may be computationally expensive when dealing with large-scale instances \cite{baldacci2011exact}. Heuristic methods can provide satisfying solutions within a reasonable time and have been widely applied in CLRP \cite{jarboui2013variable}. However, these heuristic methods always rely on handcrafted rules or domain knowledge. Recently, deep reinforcement learning (DRL) has attracted significant attention for its ability to automatically learn a policy for solving FLPs \cite{miao2024deep, liang2024sponet,zhong2024recovnet} and VRPs \cite{kwon2020pomo,wang2024deep,li2024multi}. Although these DRL-based methods have achieved promising results, they are generally limited to handling either the FLP or the VRP in isolation, rather than directly addressing the CLRP. For example, Wang \textit{et al.} \cite{wang2023new} first attempted to apply DRL to solve the CLRP by separating the original problem into two sequentially solved sub-problems. Since the location decisions and routing decisions are heavily linked to each other in CLRP \cite{salhi1989effect,salhi1999consistency}, solving sequentially may lead to inferior solution quality, which ignores the interdependencies between various decisions.

To fill this research gap, we propose the DRL with heterogeneous query (DRLHQ) to solve the CLRP. Our method follows the encoder-decoder structure \cite{vaswani2017attention} to solve the CLRP in an end-to-end manner, which is a non-trivial task to our knowledge. We first reformulate the CLRP into a markov decision process (MDP) tailored to various decisions, which is a general modeling framework that can be easily adapted to other DRL-based methods. In the CLRP, routing decisions operate at a point-to-point level within a subtour, whereas location decisions are made at a tour-to-tour level, where selecting a depot determines the global region of the next subtour. Accordingly, we propose a novel state-conditioned heterogeneous querying attention mechanism, which allows the construction of decision-specific queries at various decision stages, thereby effectively capturing the interdependencies across distinct decision stages. Meanwhile, we design a dynamic masking mechanism based on the transition rules of MDP to ensure feasibility and the policy is optimized by the REINFORCE algorithm. Extensive experimental results on synthetic and benchmark datasets show that DRLHQ outperforms other baselines, demonstrating superior solution quality and better generalization performance.

The rest of the paper is organized as follows. We briefly review the related works in Section \ref{RW}. Section \ref{PS} illustrates the mathematical formulation of CLRP. Our methodology is explained in detail in Section \ref{Methodology}. Section \ref{exp} presents simulations and experimental results. Section \ref{Conclusion} concludes this paper and discusses future work.

\section{Background and Motivation}
\label{RW}
\noindent In this section, we review the related works on the CLRP from the perspective of exact and heuristic methods and learning-based methods. Meanwhile, we illustrate the motivation of our work.

\subsection{Exact and Heuristic Methods}
The CLRP deals with the combination of facility location and vehicle routing, which involve complex constraints and dynamic characteristics. For small-scale CLRP, many studies developed exact methods based on modeling techniques and the branch-and-bound framework \cite{lawler1966branch}. Laporte \textit{et al.} \cite{laporte1981exact} first established an integer programming model to solve single-facility LRP through the branching delimitation method, which was an early attempt at solving the CLRP with exact methods. Inspired by this, many scholars struggled to design more efficient techniques by applying column generation \cite{berger2007location,akca2009branch} and cutting plane \cite{belenguer2011branch,contardo2014exact}. However, due to the NP-hard nature of CLRP, the performance of exact methods significantly decreases when dealing with large-scale CLRP. 

Therefore, more studies began to focus on heuristic methods, which can provide satisfying solutions within an acceptable time. Prins \textit{et al.} \cite{prins2006solving} proposed a two-stage heuristic method called GRASP, which was enhanced by learning components and path relinking. Based on GRASP, Duhamel \textit{et al.} \cite{duhamel2010grasp} further combined the GRASP with evolutionary local search (ELS) to improve the performance. In addition, a series of heuristic methods are also widely applied to solve the CLRP, such as iterated local search (ILS) \cite{derbel2010iterated}, simulated annealing heuristics (SAH) \cite{vincent2010simulated,vincent2015simulated}, and tree-based search algorithm (TBSA) \cite{schneider2019large}. In addition, there are also some heuristic works \cite{wang2019multiobjective} on the variant of CLRP that consider multiple objectives. Although heuristic methods have achieved some success in solving the CLRP, they heavily depend on expert domain knowledge and intricate handcrafted techniques.

\subsection{Learning-based Methods}
With the emergence of artificial intelligence, DRL has been widely applied in solving COPs, while most studies focus on routing problems. Vinyals \textit{et al.} \cite{vinyals2015pointer} proposed the Pointer Network, the first attempt to apply deep learning in solving routing problems. Inspired by the structure of transformer \cite{vaswani2017attention}, Kool \textit{et al.} \cite{kool2018attention} proposed an attention model (AM) to generate solutions in a construction manner, which laid a solid foundation for further studies. Meanwhile, Kwon \textit{et al.} \cite{kwon2020pomo} enhanced the efficiency of AM by introducing policy optimization with multiple optima (POMO). It is worth mentioning that POMO is one of the most representative and state-of-the-art DRL-based methods, which is widely utilized as the backbone. For example, Li \textit{et al.} \cite{li2024multi} proposed a multi-type attention encoder for multi-depot VRP, thereby better handling the characteristics of nodes. Furthermore, Wang \textit{et al.} \cite{wang2024deep} designed a two-stage attention-based encoder for VRP with backhauls, which can yield more informative representations. Except for routing problems, some studies have also explored the application of DRL in solving FLPs \cite{miao2024deep,liang2024sponet,zhong2024recovnet}.

Benefiting from the flexibility and strength of DRL, its application in solving VRPs and FLPs has shown promising results and remarkable efficiency. However, due to complex constraints and characteristics of CLRP, studies related to CLRP remain lacking. Li \textit{et al.} \cite{li2022heuristic} explored a heuristic approach using a Hopfield neural network to optimize the sequencing of location and routing decisions in complex production environments. Zou \textit{et al.} \cite{zou2024reinforcement} further combined reinforcement learning with evolutionary algorithms, guiding the search process for the Latency Location-Routing Problem, where minimizing latency plays a crucial role. Meanwhile, Kaleem \textit{et al.} \cite{kaleem2024neural} considered routing decisions as a surrogate model and embedded neural networks into the optimization framework to handle both location and routing tasks simultaneously. In the context of multi-echelon problems, Huang \textit{et al.} \cite{huang2025deep} designed a two-stage attention model for the Two-Echelon Location-Routing Problem, which optimizes location and routing decisions across two hierarchical levels.

Since the aforementioned studies primarily focus on specialized or hybrid methods for intricate variants of CLRP, the work most closely related to ours is by Wang \textit{et al.} \cite{wang2023new}, who proposed AM-W, a pioneering attempt to apply DRL to solve the CLRP, laying the groundwork for further exploration in this domain. They divide the CLRP into two separate decision problems and train a separate DRL model for each problem. During inference, the AM-W first calls the location model to decide which depots to open, then calls the routing model to make routing decisions based on the opened depots. Since the location decisions and routing decisions are heavily linked to each other in CLRP \cite{salhi1989effect,salhi1999consistency}, solving sequentially may lead to inferior solution quality, which ignores the interdependencies between various decisions.

\begin{figure*}[!t]
	\includegraphics[width=\textwidth]{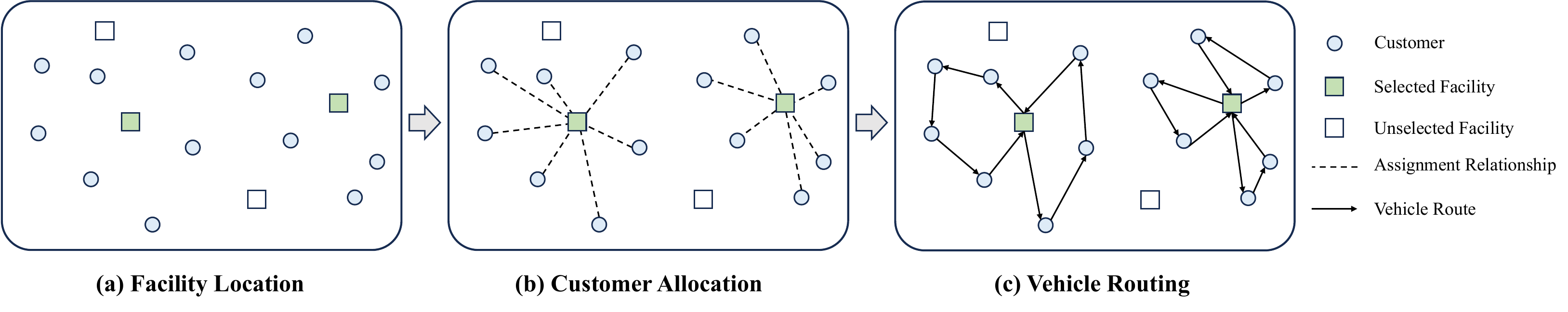}
	\caption{An illustrative example of CLRP. The decision process of CLRP can be divided into three partitions: (a) Facility Location, (b) Customer Allocation, and (c) Vehicle Routing. The decisions across different partitions are highly interdependent and strongly coupled. Each depot and vehicle is subject to capacity constraints, which makes it challenging to solve the CLRP.}
	\label{problem}
\end{figure*}

\subsection{Motivation}

Although AM-W, as the first work to apply DRL-based methods to solving the CLRP, achieved certain success, it still faces the following limitations: (1) First, the AM-W adopts a greedy approach to assign depots to customers during the location phase, and incorporating the distance between depots and customers directly into the optimization objective, which overlooks critical routing information; (2) Next, since the depot-customer assignments are completed during the location phase, the resulting allocations may produce scattered customer nodes for each depot, which require additional vehicles to service, thereby increasing total costs; (3) Finally, the location decisions partition the problem into multiple VRP instances of varying sizes, and this variation in problem scale poses significant challenges for the generalization ability of DRL-based methods.

Motivated by the above literature review, we propose an end-to-end learning approach for solving the CLRP. Specifically, we reformulate the CLRP as a unified MDP that accommodates heterogeneous decision types within a single solution construction procedure, providing a general modeling framework that can be readily extended to other DRL-based methods. Building upon this MDP formulation, we introduce a dynamic masking mechanism that is coupled with the MDP transition rules, enabling adaptive switching between decision stages and guaranteeing feasibility throughout the solution construction process. Considering that different decision stages depend on information at different levels of granularity, we propose a state-conditioned heterogeneous querying attention mechanism that constructs decision-specific queries aligned with the characteristics of each decision stage. Our approach follows an encoder-decoder architecture to automatically learn the solution construction policy in an end-to-end manner. It is worth noting that our approach is the first to address the CLRP in an end-to-end manner, rather than treating the location and routing decisions independently.

\section{Problem Statement}
\label{PS}
\noindent In this section, we introduce the mathematical formulation of CLRP. As Fig. \ref{problem} depicted, the CLRP is a classical and challenging combinatorial problem consisting of three decision-making problems: facility location, customer allocation, and vehicle routing. The decisions across different partitions are highly interdependent and strongly coupled.

The CLRP can be defined as a complete, weighted, and undirected graph $G=(V, E)$. Specifically, $V$ represents the set of nodes, which includes the subset $I$ of potential depot locations and the subset $J$ of customers. $E$ denotes the set of undirected edges connecting the nodes, and each edge $(i,j)\in E$ is associated with a positive cost $c_{ij}=c_{ji}\textgreater 0$. The subset $K$ of vehicles comprises homogeneous vehicles, each characterized by a loading capacity $q \textgreater 0$ and incurring a fixed operating cost $F$. The set $K$ is assumed to be unlimited, and each edge $e\in E$ satisfies the triangle inequality. Each customer $j\in J$ has a deterministic demand $D_j$, known in advance. For CLRP, each depot $i\in I$ has a limited capacity $Q_i$ and an opening cost $O_i$. Each customer $j\in J$ must be served by a single vehicle at exactly once, and the capacity of depots and vehicles may not be exceeded. The CLRP model was first proposed by Cooper \textit{et al.} \cite{cooper1972transportation}, and the CLRP can be formulated as follows:

\begin{small}
%\begin{subequations}
%\label{model}
\begin{align}
&\min \quad \sum_{i \in I} O_i y_i+\sum_{i \in V} \sum_{j \in V} \sum_{k \in K} c_{i j} x_{i j k}+\sum_{i \in I} \sum_{j \in J} \sum_{k \in K} F x_{i j k}\label{Obj}\\
&\text{subject to}\notag\\
&\sum_{i \in V} \sum_{k \in K} x_{i j k}=1 \quad \forall j \in J \label{2}\\
&\sum_{i \in I} \sum_{j \in J} x_{i j k} \leq 1 \quad \forall k \in K \label{3}\\
&\sum_{j \in V} x_{i j k}-\sum_{j \in V} x_{j i k}=0 \quad \forall k \in K, \forall i \in V \label{4}\\
&U_{ik} +D_i - U_{jk} \leq q \cdot (1-x_{ijk}) \quad \forall i,j \in J, i \neq j, \forall k \in K  \label{5}\\
&\sum_{u \in J} x_{i u k}+\sum_{u \in V \backslash\{j\}} x_{u j k} \leq 1+z_{i j} \quad \forall i \in I,  \forall j \in J, \forall k \in K  \label{6}\\
&\sum_{i \in V} \sum_{j \in J} D_j x_{i j k} \leq q \quad \forall k \in K \label{7}\\
&\sum_{j \in J} D_j z_{i j} \leq Q_i y_i \quad \forall i \in I \label{8}\\
&\sum_{i \in I} \sum_{j \in J} \sum_{k \in K} x_{ijk} \leq |K| \label{9}\\
&x_{ijk} = 0\quad \forall i\in I,\forall j \in I,\forall k \in K \label{10}\\
&x_{ijk}\in \{0,1\}\quad \forall i \in V,\forall j \in V,\forall k \in K \label{11}\\
&y_i\in\{0,1\}\quad\forall i \in I \label{12}\\
&z_{ij}\in\{0,1\}\quad\forall i \in I,\forall j \in V \label{13}\\
&U_{ik }\geq 0\quad \forall i \in V,\forall k \in K \label{14}
%&c_{ji}=0\quad \forall i \in I,\forall j \in J \label{15}
\end{align}
%\end{subequations}
\end{small}

\begin{figure*}
	\centering
	\includegraphics[width=\textwidth]{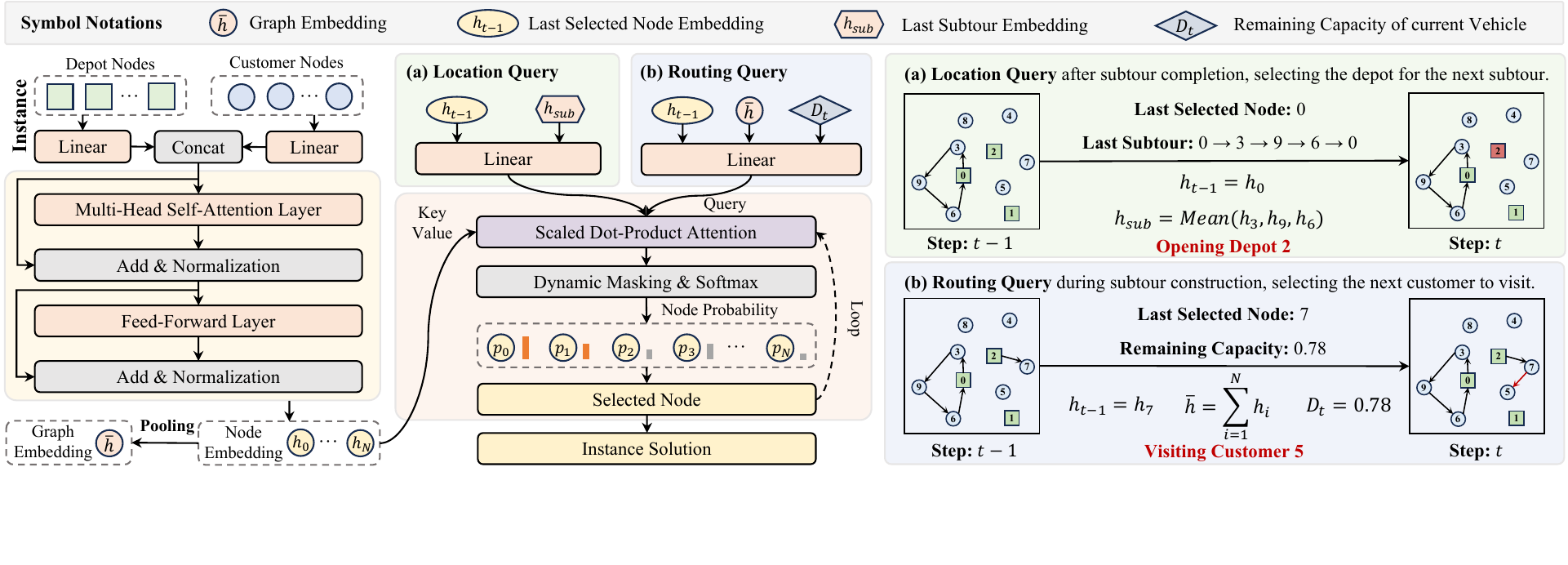}\vspace{-1cm}
	\caption{The overall pipeline of DRLHQ. We propose a state-conditioned heterogeneous querying attention mechanism that constructs decision-specific query vectors tailored to different decision stages: (a) after completing a subtour, a location query is constructed to select the starting depot for the next subtour, conditioned on the embedding of the last selected node and an aggregated representation of the previous subtour; (b) during subtour construction, a routing query is constructed to select the next customer within the current subtour, based on the embedding of the last selected node, the global graph embedding, and the remaining vehicle capacity.}
	\label{pipeline}
\end{figure*}

The objective function (\ref{Obj}) aims to minimize the total cost, including the opening cost of depots, the traveling cost of routes, and the fixed cost of vehicle utilization. Constraint (\ref{2}) ensures that each customer is visited by a single vehicle exactly once. Constraints (\ref{3}) and (\ref{4}) jointly guarantee that each vehicle departs from at most one depot and returns to the same depot after completing all assigned services. Constraint (\ref{4}) also ensures the continuity of each route, which makes each vehicle depart from the customer after service. Constraint (\ref{5}) indicates the subtour elimination constraints. Constraint (\ref{6}) specifies that a customer can only be allocated to a depot if an available route exists. Constraints (\ref{7}) and (\ref{8}) ensure that the capacity of each vehicle and the capacity of each depot is not exceeded. Constraint (\ref{9}) limits the number of vehicles used. Constraint (\ref{10}) prevents the connections between depots. Constraints (\ref{11}) to (\ref{14}) define the decision variables. The related notations are listed as follows:

\noindent Sets

\begin{itemize}
	\item $V$: Set of all nodes, represented as $V=I \bigcup  J$.
	\item $I$: Set of potential depot locations.
	\item $J$: Set of customers requiring service.
	\item $E$: Set of edges connecting all nodes.
	\item $K$: Set of vehicles.
\end{itemize}
 
\noindent Parameters

\begin{itemize}
	\item $O_i$: Fixed cost of opening a depot at node $i$.
	\item $Q_i$: Capacity of the depot located at node $i$.
	\item $D_j$: Demand of the customer located at node $j$.
	\item $q$: Loading capacity per vehicle.
	\item $F$: Fixed cost of operating a vehicle.
	\item $c_{ij}$: Traveling cost associated with the edge $(i,j)$.
\end{itemize}

\noindent Decision Variables

\begin{itemize}
	\item $x_{ijk}$: is equal to 1 if the vehicle $k$ traverses the edge from customer $i$ to customer $j$, and 0 otherwise.
	\item $y_i$: is equal to 1 if the depot located at node $i$ is opened, and 0 otherwise.
	\item $U_{ik}$: the cumulative load of vehicle $k$ at node $i$.
	\item $z_{ij}$: is equal to 1 if customer $j$ is served by depot $i$, and 0 otherwise.
\end{itemize}

\section{Methodology}
\label{Methodology}

\noindent In this section, we present the details of DRLHQ, an end-to-end learning approach for solving the CLRP. We first reformulate the problem as a unified Markov decision process (MDP). Based on this formulation, our method adopts an encoder-decoder architecture \cite{vaswani2017attention} to parameterize the policy. To accommodate heterogeneous decision requirements during solution construction, including facility location and vehicle routing, we propose a state-conditioned heterogeneous querying attention mechanism that constructs decision-specific queries. Finally, the policy is optimized using the REINFORCE algorithm \cite{williams1992simple}, and the overall pipeline of DRLHQ is illustrated in Fig.~\ref{pipeline}.

\subsection{Markov Decision Process Formulation}
\label{mdp}
It is essential to reformulate the problem into an MDP to solve it by DRL. Since customers can only be served by opened depots, the CLRP must make location decisions before routing decisions. However, the solution construction of CLRP can also be considered a sequential decision-making process \cite{keneshloo2019deep} with the help of proper reformulation. The construction of a solution can be equivalently decomposed into the construction of several subtours. Each subtour is composed of one depot and several customers. Therefore, the location decision is implicitly embedded within the solution representation, i.e., a depot is considered unopened if it is not included in the solution. The reformulation can be easily extended to other DRL-based methods.

In particular, the MDP can be defined as a five-tuple $\{S, A, T, R, P\}$, where $S$ denotes the state space, $A$ denotes the action space, $T$ denotes the state transition, $R$ denotes the rewards, and $P$ denotes the policy. Assuming there are $|I|$ potential depots and $|J|$ customers, where the sets $I$ and $J$ are mutually independent.

\emph{\textbf{State:}} The state $S^t=\{P^t,V^t,M^t,X^t,I^t\}$ consists of five components. The first component $P^t$ indicates the current partial solution, i.e., $P^t=\{x^0,x^1,...,x^t\}$ is the sequence of visited nodes sequence till step $t$, where $x^i$ is the node visited at step $i$. The second component $V^t=\{C_V^t, L^t\}$ denotes the vehicle state, where $C_V^t$ is the remaining capacity of the vehicle at step $t$, $L^t$ is the total routing length till step $t$. The third component $M^t=\{x_i^t|i=1,2,...,|I|\}$ is related to the depot state, i.e., $x_i^t=(G_i,F_i,C_i^t)$, where $G_i$ denotes the two-dimensional coordinates, $F_i$ is the fixed cost of opening the depot, and $C_i^t$ is the remaining capacity of the depot. The component $X^t=\{x_j^t|j=|I|+1,|I|+2,...,|I|+|J|\}$ is related to the customer state, i.e., $x_j^t=(G_j,D_j^t)$, where $G_j$ is the two-dimensional coordinates, and $D_j^t$ is the demand of the customer at step $t$. Moreover, the last component $I^t\in\{0,1\}$ is the indicator state of decisions at step $t$, leading to various action spaces. Here, $I^t=1$ indicates that a subtour has concluded, while $I^t = 0$ denotes that the subtour is ongoing.

\emph{\textbf{Action:}} The action $A\in \{A_c \cup A_d\}$ denotes the action space at step $t$, which can be decomposed into two partitions: (1) The routing action $A_c$ is defined as selecting the next customer to be visited when indicator state $I^t=0$; (2) The location action $A_d$ involves selecting the starting depot for next subtour when indicator state $I^t=1$, regardless of whether the depot is closed or opened. It should be noted that the action space at step $t$ will adaptively switch according to the indicator state $I^t$. To ensure the feasibility of the solution, all nodes that violate constraints are dynamically masked at each step.

\emph{\textbf{Transition:}} The state $S^t$ transits to the next state $S^{t+1}$ after taking the action $A^t\in A$, which follows the transition rules. Assuming that the node selected at step $t+1$ is $x_j$, the selected node is directly added to the partial solution $P^{t+1}=[P^t,x_j]$. The remaining capacity of the vehicle $C_V^t$ updates following the transition rules:
		\begin{itemize}
			\item When the selected node $x_j$ is a customer, the vehicle’s remaining capacity is reduced by the demand $D_j$. 
			\item When the selected node $x_j$ is a depot, the vehicle’s remaining capacity is updated to the minimum of the depot’s remaining allocable capacity and the vehicle’s maximum capacity.
		\end{itemize}

It should be noted that each depot may include multiple subtours, with each subtour corresponding to a separate vehicle. The vehicle state $V^t=\{C_V^t,L^t\}$ is updated as follows:

\begin{small}
\begin{equation}
\label{vehicle}
C_V^{t+1}=\left\{
\begin{aligned}
&C_V^t - D_j&, & \text{if $x_j$ is not a depot} \\
&\min(C_{max},C_j^t)&, & \text{if $x_j$ is a depot}
\end{aligned}
\right.
\end{equation}
\end{small}

\begin{small}
\begin{equation}
L^{t+1}=L^t+Z(x^t,x^{t+1})
\end{equation}
\end{small}	
where $C_{max}$ is the vehicle's maximum capacity, $C_j^t$ is the remaining allocable capacity of the visiting depot, $Z$ indicates the distance between two nodes, and the distance between two arbitrary depots is set to 0. Assuming that the departure depot of the current subtour is $x_i$, the remaining capacity of the depot $x_i$ is updated as follows:

\begin{small}
\begin{equation}
\label{depot}
C_i^{t+1}=\left\{
\begin{aligned}
& C_i^t - D_j&, & \text{if $x_j$ is not a depot} \\
& C_i^t&, & \text{if $x_j$ is a depot}
\end{aligned}
\right.
\end{equation}
\end{small}
where $x_j$ is the selected node at step $t+1$. Since customer $x_j$ is satisfied after visiting, the demand $D_j^t$ will be updated as $D_j^{t+1}=0$. 

\emph{\textbf{Reward:}} After the state terminates, we can calculate the total cost according to the Eq. \ref{Obj}. To minimize the total cost, we directly take the negative value of the objective function as the reward $R$. It should be noted that previous method \cite{wang2023new} typically solves the CLRP by first deciding which depots to open and then performing routing based on the opened depots. In contrast, our approach integrates depot selection and routing decisions into a unified MDP framework, allowing both to be considered simultaneously. As a result, the status of each depot is determined only after the solution construction is completed. Specifically, a depot is considered open if it is selected and included in the final solution; otherwise, it is regarded as closed.

\emph{\textbf{Policy:}} The goal of DRL is to find a policy $\pi_\theta$ parameterized by $\theta$, which starts from the original state $S^0$ and ends at the terminal state $S^T$. The entire process is constructed through a node-by-node selection method. The joint probability distribution of the process can be expressed as:

\begin{small}
\begin{equation}
p(S^T|S^0)=\prod_{t=0}^{t-1}\pi_\theta(A^t|S^t)
\end{equation}
\end{small}
where $\pi_\theta(\cdot|S^t)$ indicates the probability distribution at step $t$.

\begin{figure*}[!t]
	\centering
	\includegraphics[width=\textwidth]{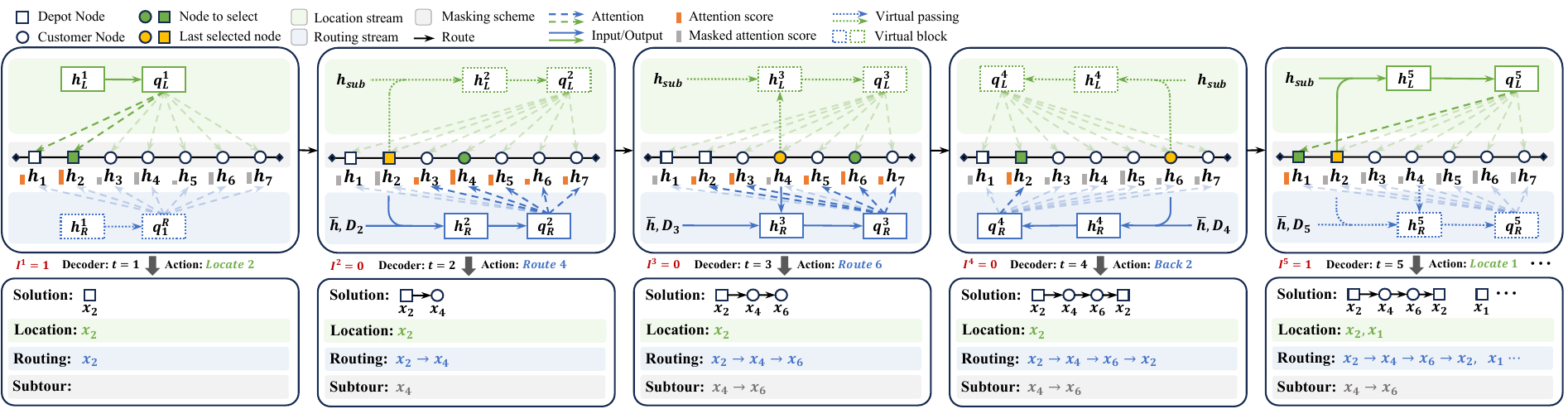}
	\caption{Illustration of the state-conditioned heterogeneous querying attention mechanism on an instance with two depots and five customers. The upper part visualizes how decision-specific queries are computed, while the lower part shows how the solution is constructed based on the selected node. At each step $t$, the mechanism computes the attention score to determine which node to select. Specifically, it maintains a location query $q_L$ and a routing query $q_r$, and adaptively switches between them according to the current decision stage indicator $I^t$: when $I^t=1$, the model performs a location decision based on $q_L$; otherwise, it performs a routing decision using $q_R$.}
	\label{decoder}
\end{figure*}
\subsection{Encoder}

The encoder takes node features as input and produces embeddings that are subsequently used by the decoder. Compared to conventional VRP, the CLRP additionally involves location decisions that must account for depot opening costs and maximum capacities. Accordingly, depot representations incorporate not only two-dimensional geographical coordinates but also the attributes related to capacity and cost. Since the scales of depot opening costs and capacities vary across instances, these attributes are normalized for consistent encoding. In particular, two ratio-based features are used when constructing depot representations: the ratio of depot capacity to opening cost, which reflects the supply capacity per unit cost, and the ratio of depot capacity to the total customer demand, which reflects the supply capability of a depot relative to the overall customers.

Each depot is thus represented by a 4-dimensional feature vector consisting of its 2D coordinates and two normalized attributes, whereas each customer is represented by a 3-dimensional feature vector consisting of its 2D coordinates and demand. As depots and customers have feature vectors of different dimensionalities, we project them into a shared $d_h$-dimensional embedding space via separate linear transformations, yielding the initial embeddings $h^{(0)}$. To capture the interdependencies between nodes, we apply the self-attention mechanism \cite{vaswani2017attention} to obtain the node embeddings. Specifically, the initial embeddings are passed through $L$ attention layers, and the output embeddings of the $l$-th layer $h^l$ are given by:

\begin{small}
\begin{align}
	&\hat{h}=\texttt{Norm}(h^{(l-1)}+\texttt{MHA}^l(h^{(l-1)}))\\
	&h^l=\texttt{Norm}(\hat{h}+\texttt{FF}(\hat{h}))
\end{align}	
\end{small}
where $\texttt{Norm}$, $\texttt{MHA}$, and $\texttt{FF}$ denote the normalization layer, multi-head attention layer, and feed-forward layer. The details of these operations are similar to the study \cite{kool2018attention}, and the node embedding passing through $L$ layers is denoted as $h_i=h_i^L$. The encoder also calculates the graph embedding of the instance, which is defined as the mean of all node embeddings:

\begin{small}
\begin{equation}
	\overline{h}=\frac{1}{N}\sum_{i=1}^N h_i
\end{equation}	
\end{small}
where $N$ is the total number of nodes, and both node embedding $h_i$ and graph embedding $\overline{h}$ are taken as the input of the decoder.

\subsection{Decoder}

Benefiting from the MDP formulation, the decoder incrementally constructs a solution based on the embeddings produced by the encoder. The solution is composed of multiple subtours, each corresponding to a vehicle route originating from a selected depot. During solution construction, two types of decisions with different granularities are involved. Location decisions are made at the tour level to select the starting depot for a new subtour, while routing decisions are made at the node level to determine the next customer to visit within an ongoing subtour. Motivated by this distinction, we propose a state-conditioned heterogeneous querying attention mechanism that constructs decision-specific queries to coordinate location and routing decisions during solution construction, as illustrated in Fig. \ref{decoder}.

\emph{\textbf{Location Query:}} 
Intuitively, location decisions determine the starting depot of the next subtour and thus specify where the subsequent subtour is positioned in the global region. Consequently, effective location decisions should leverage the state information of the previously completed subtour. To this end, we construct the context embedding for location decisions $h_t^{L}$ by concatenating the embedding of the last visited node $h_{\pi_{t-1}}$ (i.e., the depot associated with the last subtour in location decision) with an aggregated representation of the last subtour $h_{sub}$, which is defined as:

\begin{small}
\begin{equation}
	h_t^{L}=[h_{\pi_{t-1}},h_{sub}] 
\end{equation}	
\end{small}
where the last subtour embedding $h_{sub}$ is obtained by mean aggregation over the embeddings of all customers visited in the most recently completed subtour, excluding the depot. Specifically, it is computed as:

\begin{small}
\begin{equation}
	h_{sub} = \frac{1}{|\mathcal{S}_{t-1}|} \sum_{i \in \mathcal{S}_{t-1}} h_i
\end{equation}
\end{small}
where $\mathcal{S}_{t-1}$ denotes the set of customers visited in the last subtour. Accordingly, the location query at step $t$ is obtained by applying a linear projection to the location context embedding, given by:
\begin{small}
\begin{equation}
	q_t^{L} = h_t^{L} W_q^{L} .
\end{equation}
\end{small}
where $W_q^{L}$ denotes the linear projection for constructing the location query.

\emph{\textbf{Routing Query:}}  
In contrast, routing decisions determine the next customer to be visited during subtour construction, thereby shaping the local structure of the route. As routing decisions operate at the node level and are closely coupled with the current vehicle state, they should incorporate information from the ongoing subtour to support customer selection. Therefore, we construct the context embedding for routing decisions $h_t^{R}$ by concatenating the embedding of the last visited node $h_{\pi_{t-1}}$, the graph embedding $\overline{h}$, and the remaining capacity of current vehicle $D_t$, which is defined as:

\begin{small}
\begin{equation}
	h_t^{R}=[h_{\pi_{t-1}}, \overline{h}, D_t]
\end{equation}	
\end{small}
Based on the routing context embedding, the routing query at step $t$ is obtained by applying a linear projection, given by:

\begin{small}
\begin{equation}
	q_t^{R} = h_t^{R} W_q^{R}
\end{equation}
\end{small}
where $W_q^{R}$ denotes the linear projection for constructing the routing query.

\emph{\textbf{Decoding:}}  
Since the construction of the \emph{query} dynamically changes according to various decision stages, we can manage the \emph{query} at each step using the following formula: 

\begin{small}
\begin{equation}
	Q^t=q_t^{L}I^t + q_t^{R}(1-I^t)
\end{equation}	
\end{small}
where $I^t$ denotes the indicator state at step $t$, which enables the model to adaptively select the appropriate query according to the current decision stage. The decoder also yields the \emph{key} and \emph{value} from the node embeddings through linear projections, which can be expressed as follows:

\begin{small}
\begin{equation}
	K_i=h_iW_k,\quad V_i=h_iW_v
\end{equation}	
\end{small}
where $K_i$ and $V_i$ denote the $key$ and $value$ of the $i$-th node, $h_i$ is the context embedding, and $W_k$ and $W_v$ are learnable parameters. Similar to the study \cite{vaswani2017attention}, we compute the attention through a multi-head attention mechanism, which can be computed as follows:

\begin{small}
\begin{equation}
	u_i^t=C\cdot tanh(\frac{Q_m^tK_m}{\sqrt{d_{K}}})V_m
\end{equation}	
\end{small}
where $C$ is a constant hyper-parameter, $tanh$ is an activation function, $m=1,2,...,M$ denotes the head in the mechanism ($M=16$ is the number of heads), and $d_{K}=\frac{d_h}{M}$ is the scaling factor. To ensure the feasibility, we set a large negative value to the nodes violating the constraints, and the node selection probability for selecting the $i$-th node can be derived by applying a Softmax function as follows:

\begin{small}
\begin{equation}
	p_i = softmax(\overline{u}_i^t)=\frac{exp(\overline{u}_i^t)}{\sum_{i=1}^nexp(\overline{u}_i^t)}
\end{equation}	
\end{small}
where $p_i$ denotes the node selection probability and $\overline{u}_i^t$ is the attention after applying the mask. Based on the node selection probabilities provided by the decoder, a solution can be constructed by applying proper search strategies.

\subsection{Dynamic Masking Mechanism}

Unlike previous methods that first determine depot selection and then perform routing, our method integrates depot selection and routing decisions during the solution construction process. To ensure the feasibility of the obtained solutions, we propose a dynamic masking mechanism to handle the complex constraints and intricate relationships between various decisions in LRPs. Specifically, each decision stage corresponds to distinct masking rules during the decoding process. To manage heterogeneous queries and regulate transitions between different decision stages, we introduce an indicator state $I^t$ within the MDP formulation, which explicitly encodes the current decision stage.

To handle the complex constraints and ensure solution feasibility, our dynamic masking mechanism defines distinct masking rules for each indicator state.

\subsubsection{Indicator State $I^t=0$}
In this state, the model can either selects the unmasked customer or returns to the departure depot, which should adopt the following masking rules:  

\begin{theorem}
	\label{R1}
	All previously selected customers should be masked.
\end{theorem}

\begin{theorem}
	\label{R2}
	All depots should be masked when $I^t=0$, except for the departure depot of the current subtour.
\end{theorem}

\begin{theorem}
	\label{R3}
	Customers whose demand exceeds the remaining capacity of the current vehicle or the remaining capacity of the starting depot are masked.
\end{theorem}

\begin{theorem}
	\label{R4}
	If the process is at the beginning of a subtour, having just selected the departure depot but not yet selected any customer, the departure depot should be masked.
\end{theorem}

Masking Rule \ref{R1} ensures that each customer is selected only once, explicitly satisfying Constraint (\ref{2}). Masking Rule \ref{R2} ensures the vehicle must return the departure depot, which is described in Constraint (\ref{4}). Masking Rule \ref{R3} guarantees that each selected node during route construction must comply with the capacity limits of both the depot and the vehicle, explicitly satisfying Constraints (\ref{7}) and (\ref{8}). After the model selects a node to visit, the remaining capacities of the vehicle and the depot are immediately updated according to Equations (\ref{vehicle}) and (\ref{depot}). Masking Rule \ref{R4} is proposed to prevent the connections between depots, satisfying the Constraint (\ref{10}).

\subsubsection{Indicator State $I^t=1$}
Similarly, when a subtour is completed, the model must select the departure depot for the next subtour, which should follow the masking rule:

\begin{theorem}
	\label{R5}
	A depot should be masked if its remaining capacity cannot satisfy the minimum demand among the unselected customers.
\end{theorem}
 
Masking Rule \ref{R5} prevents the selection of depots with insufficient capacity, which could lead to failure in constructing subsequent subtours. It ensures the proper construction of subtours under Constraints (\ref{3}) and (\ref{4}). Constraints (\ref{5}) and (\ref{6}), which address subtour elimination and assignment respectively, are implicitly satisfied through the MDP formulation and the solution construction process. As for Constraint (\ref{9}), since the maximum number of vehicles is defined as the number of customers, the constraint is inherently satisfied.

By applying the dynamic masking mechanism, the model can simultaneously consider depot selection and routing decisions and switch between them flexibly, without relying on a fixed decision order. Additionally, the dynamic masking mechanism helps the model capture the intrinsic correlations between decisions, avoiding performance degradation caused by considering	 them separately.

\subsection{Model Training}
Our method takes POMO \cite{kwon2020pomo} as the backbone algorithm, which proposes a multi-trajectories training framework to improve performance. Similar to POMO, we sample a set of trajectories $\{\tau^1,\tau^2,...,\tau^N\}$ by selecting various starting nodes for each instance. To maximize the total reward $J$, the gradient used for gradient descent can be computed with an approximation as follows:

\begin{small}
\begin{equation}
	\nabla_\theta J(\theta)\approx\frac{1}{N}\sum_{i=1}^N(R(\tau^i)-b^i(s))\nabla_\theta \log p_\theta(\tau^i|s)
\end{equation}	
\end{small}
where $R(\tau^i)$ and $p_\theta(\tau^i|s)$ are the total reward and joint probability distribution of the trajectory $\tau^i$. Specifically, $b^i(s)$ is a shared baseline computed by the mean of all trajectories:

\begin{small}
\begin{equation}
	b^i(s)=b_{shared}=\frac{1}{N}\sum^N_{j=1}R(\tau^j),\quad \text{for all $i$}
\end{equation}	
\end{small}

Therefore, the policy can be optimized by the REINFORCE algorithm \cite{williams1992simple} based on a shared baseline.

\section{Experimental Results}
\label{exp}
\noindent In this section, we conduct comprehensive experiments to evaluate the effectiveness of DRLHQ. We first detail the experimental settings and then compare our method against other representative baselines using synthetic datasets. Additionally, we evaluated the generalization performance of our method on cross-scale and cross-distribution cases, respectively.

\subsection{Experimental Settings}
\subsubsection{Datasets}
Our experiments are primarily conducted on two types of datasets: synthetic and benchmark datasets. For the synthetic dataset, we keep the consistency with the related works \cite{kool2018attention,kwon2020pomo,li2024multi,wang2024deep}, generating depot and customer nodes by uniformly sampling within a unit square $[0\times1]^2$. We consider four problem scales with the number of customers $N=$ 10, 20, 50, and 100, where each problem scale contains 5, 5, 10, and 20 potential depots, respectively. For each problem scale, we generate 1,000 independent instances to ensure statistically reliable evaluation. Since nodes are not uniformly distributed in real-world scenarios, we conduct comparative experiments on publicly available real datasets. Specifically, we employ the instances designed by Prins \textit{et al.} \cite{prins2004nouveaux} as the benchmark dataset, which has a more practical node distribution and quantity of demands. 

\subsubsection{Training}
All hyperparameters related to training and model architecture are kept the same with the related works \cite{kool2018attention,kwon2020pomo,li2024multi,wang2024deep}. Specifically, the embedding dimension is set to 256, the number of attention layers is 6, the number of heads in multi-head attention is 16, and the dimension of the feed-forward layer is 512. As for the training setup, we train the model for 1,000 epochs at each problem scale, with each epoch comprising 10,000 synthetic instances. We use an initial learning rate of 1e-4 and decay it to 1e-5 after 700 epochs. The batch size is set to 128 by default, and it is reduced to 40 for the scale of 100 due to the limit of memory. All experiments are conducted on the server with a GTX 4090 GPU and Intel(R) Xeon(R) Gold 6230 CPU @ 2.10GHz.

\subsubsection{Inference}
To ensure a fair comparison with the exact solver and heuristic methods, we solve instances for all DRL-based methods individually rather than in parallel. For DRL-based methods, we report results under both greedy decoding and greedy decoding with ($8\times$) instance augmentation following POMO \cite{kwon2020pomo}. Furthermore, our DRLHQ can be seamlessly integrated with the recent simulated-guided beam search (SGBS) \cite{choo2022simulation}, resulting in a variant denoted as DRLHQ-SGBS, which achieves superior performance without requiring any additional modifications. We report the average objective value and optimality gap on each dataset, along with the average runtime per instance. Specifically, the gaps are computed relative to the best-performing method on each dataset.

\subsubsection{Baseline Algorithms}
To evaluate the effectiveness of our method, we compare DRLHQ with various types of baselines, including the exact solver, conventional heuristic methods, and other DRL-based methods. For each heuristic method, we provide two kinds of parameter configurations: a fast version and a slower one, where the slower one conducts a more thorough search. Specifically, the baseline methods are detailed as follows:

\begin{table*}[!t]
\caption{Results on Synthetic Datasets.}
\centering
\resizebox{\textwidth}{!}{
\begin{threeparttable} %添加此处	
\begin{tabular}{cc|ccc|ccc|ccc|ccc}
\toprule
&\multirow{2}{*}{Method} & \multicolumn{3}{c|}{CLRP10}                                                                                & \multicolumn{3}{c|}{CLRP20}                          & \multicolumn{3}{c|}{CLRP50}                          & \multicolumn{3}{c}{CLRP100}                          \\
                 &  & Obj.                             & Gap$^\dagger$                               & Time (s)                            & Obj.             & Gap$^\dagger$             & Time (s)        & Obj.             & Gap$^\dagger$             & Time (s)        & Obj.             & Gap$^\dagger$             & Time (s)        \\ \midrule
\multirow{2}{*}{\rotatebox[origin=c]{90}{Exact}}
&Gurobi (1800s)    & \multirow{2}{*}{\textbf{8.6007}} & \multirow{2}{*}{\textbf{0.00\%}} & \multirow{2}{*}{\textbf{1163.0850}} & 14.9815 & 1.58\%  & 1800     & 35.3001 & 30.54\% & 1800     & 247.0554 & 349.02\% & 1800     \\
&Gurobi (3600s)    &                         &                         &                           & 14.9626 & 1.45\%  & 3600     & 30.4853 & 12.73\% & 3600     & 223.1154 & 305.51\% & 3600     \\ \midrule
\multirow{5}{*}{\rotatebox[origin=c]{90}{Heuristics}}
&ILS               & 9.1049                  & 5.86\%                  & 4.6411                    & 16.2058 & 9.88\%  & 10.7528  & 32.9861 & 21.98\% & 19.9312  & 69.9136  & 27.07\%  & 30.2592  \\
&ILS$^\ddagger$              & 9.0834                  & 5.61\%                  & 47.7900                   & 16.1621 & 9.58\%  & 118.2543 & 32.0490 & 18.52\% & 233.4540 & 67.0150  & 21.80\%  & 532.3572 \\ \cmidrule{2-14}
&SAH               & 9.8376                  & 14.38\%                 & 20.8391                   & 16.2202 & 9.98\%  & 40.9745  & 31.7438 & 17.39\% & 95.1364  & 66.2166  & 20.35\%  & 199.6815 \\
&SAH$^\ddagger$              & 9.7989                  & 13.93\%                 & 44.3685                   & 16.2655 & 10.29\% & 95.2845  & 31.5091 & 16.52\% & 417.6718 & 64.7339  & 17.65\%  & 398.0102 \\ \cmidrule{2-14}
&TBSA$^\star$              & 9.1410                  & 6.28\%                  & 6.9960                    & 16.0016 & 8.50\%  & 23.3284  & 32.7263 & 21.02\% & 492.2040 & 67.0045  & 21.78\%  & 709.7547 \\ \midrule
\multirow{10}{*}{\rotatebox[origin=c]{90}{DRL-based Methods}}
&AM-W              & 9.4586                  & 9.97\%                  & 0.0469                    & 18.0674 & 22.50\% & 0.0727   & 37.2471 & 37.74\% & 0.1583   & 70.5208  & 28.17\%  & 0.2812   \\ \cmidrule{2-14}
&POMO              & 9.4551                  & 9.93\%                  & 0.0492                    & 18.1263 & 22.90\% & 0.0822   & 37.2600 & 37.79\% & 0.2136   & 70.2959  & 27.76\%  & 0.5064   \\
&POMO*-Greedy      & 8.7239                  & 1.43\%                  & 0.4317                    & 15.1470 & 2.70\%  & 0.4803   & 28.1288 & 4.02\%  & 0.6186   & 56.2868  & 2.30\%   & 0.9374   \\
&POMO*-Aug         & 8.6516                  & 0.59\%                  & 0.4849                    & 14.9092 & 1.09\%  & 0.5106   & 27.6849 & 2.38\%  & 0.6343   & 55.7038  & 1.24\%   & 0.9888   \\ \cmidrule{2-14}
&MTA               & 9.4054                  & 9.36\%                  & 0.0593                    & 18.0581 & 22.44\% & 0.0962   & 37.2247 & 37.66\% & 0.2232   & 70.5648  & 28.25\%  & 0.5898   \\
&MTA*-Greedy       & 8.7110                  & 1.28\%                  & 0.5621                    & 15.1450 & 2.69\%  & 0.5797   & 28.0739 & 3.82\%  & 0.7532   & 56.3390  & 2.39\%   & 1.2430   \\
&MTA*-Aug          & 8.6527                  & 0.60\%                  & 0.5909                    & 14.9214 & 1.17\%  & 0.6215   & 27.6102 & 2.10\%  & 0.8014   & 55.7463  & 1.32\%   & 1.2947   \\ \cmidrule{2-14}
&DRLHQ-Greedy      & 8.6988                  & 1.14\%                  & 0.4425                    & 15.0451 & 2.01\%  & 0.4581   & 27.7717 & 2.70\%  & 0.6822   & 55.9037  & 1.60\%   & 1.0498   \\
&DRLHQ-Aug         & 8.6480                  & 0.55\%                  & 0.4780                    & 14.8549 & 0.72\%  & 0.4643   & 27.3326 & 1.08\%  & 0.7078   & 55.4599  & 0.80\%   & 1.2365   \\
&DRLHQ-SGBS        & 8.6293                  & 0.33\%                  & 0.9357                    & \textbf{14.7485} & \textbf{0.00\%}  & \textbf{2.0782}   & \textbf{27.0417} & \textbf{0.00\%}  & \textbf{9.3211}   & \textbf{55.0214}  & \textbf{0.00\%}   & \textbf{33.3025} \\ \bottomrule
\end{tabular}
      \begin{tablenotes} %添加此处
      \footnotesize
		\item $\ddagger$ The methods marked with $^\ddagger$ adopt a parameter configuration with longer computation time, enabling more thorough search at the cost of increased computation time.
		\item $\star$ We follow the TBSA$_{basic}$ settings described in original paper for CLRP10 and CLRP20, and use TBSA$_{speed}$ for larger-scale problems.
		\item $\dagger$ The gap is calculated upon the best-performing method with the lowest objective value, where the best-performing method is in \textbf{bold}.
		\item * The method marked with * means that the method is adapted by the dynamic masking mechanism which is proposed in this work.
     \end{tablenotes} %添加此处
\end{threeparttable} %添加此处
}
\label{comparison1}
\end{table*}

\begin{itemize}
	\item \textbf{Gurobi} \cite{gurobi}, a commercial exact solver for solving mixed integer programming (MIP) problems. We simultaneously set a maximum time limit of 1,800 seconds and 3,600 seconds for each testing instance to examine the performance differences of Gurobi under different parameter settings.
	\item \textbf{ILS} \cite{derbel2010iterated}, a conventional heuristic method based on iterated local search. We set the maximum times of local search as 100, and the search will be terminated if there is no improvement in 100 iterations. The maximum times of local search is set to 1,000 in slower version. 
	\item \textbf{SAH} \cite{vincent2010simulated}, another representative heuristic method based on simulated annealing. In particular, the solutions that violate the depot capacity constraints will be penalized 400 per unit. Additionally, we set the number of iterations for the search process and the maximum number of local search to 1,000. As to the slower version, the two parameters are extended to 5,000 and 3,000, respectively.
	\item \textbf{TBSA} \cite{schneider2019large}, A tree-based search algorithm (TBSA) that explores the space of depot configurations in a tree-like fashion using a customized first improvement strategy. Considering the high computational complexity of TBSA for large-scale problems, and given that the original paper provides multiple parameter configurations, we adopt different configurations based on problem scale.
	\item \textbf{AM-W} \cite{wang2023new}, the first study applying DRL-based method to solve the CLRP, which divides the problem into two independent sub-problem and solves each using separate DRL models. This study makes a preliminary attempt to solve the CLRP with DRL, which helps to further highlight the contribution of our work.
	\item \textbf{POMO} \cite{kwon2020pomo}, a state-of-the-art DRL-based method for solving routing problems. It enhances the performance by exploration of multiple trajectories and instance augmentation, which is originally designed for generic routing problems. To better demonstrate the novelty of our work, we adapt POMO to CLRP by incorporating our proposed dynamic masking mechanism to coordinate depot selection and customer routing. We report results for both the vanilla POMO and its adapted variant.
	\item \textbf{MTA} \cite{li2024multi}, a strong DRL-based method developed based on POMO. It proposes multi-type attention to solve the multi-depot VRP by combining different types of embeddings. Similar to POMO, we apply our dynamic masking mechanism to adapt MTA for CLRP, yielding a variant for comparison.

\end{itemize}

\subsection{Comparison Study on Synthetic Datasets}
We first evaluate the performance on synthetic datasets, each consisting of 1,000 instances, and report all metrics averaged over the entire dataset.

\subsubsection{Performance Comparison}
 For the exact solver Gurobi, we set a maximum time limit of 1,800 for each instance. The results show that it takes nearly 20 minutes to solve a single instance involving 10 customer nodes to the optimal. As the problem scale increases, Gurobi can only provide suboptimal solutions within the time limit. To thoroughly evaluate the performance of Gurobi, we extend the maximum time limit from 1,800 to 3,600 seconds, while the performance improvement remains marginal. Specifically, the performance gap between Gurobi and our method increases to 12.73\% on CLRP50 and reaches 305.51\% on CLRP100. Regarding heuristic baselines, ILS, SAH, and TBSA can provide feasible solutions across various problem scales. However, even for small problem scales, the minimum performance gap for these heuristics remains at least 5.61\%. We also compare our DRLHQ against other representative DRL-based baselines using various decoding strategies. Our method consistently outperforms other DRL-based baselines across all cases while maintaining comparable computational times. Specifically, by incorporating SGBS, our method is able to flexibly trade computational time for improved solution quality, achieving substantially better performance with a moderate and acceptable increase in inference time.

\subsubsection{Efficiency Comparison}
In terms of computation efficiency, all methods experience increased computational times as the problem scale grows, especially for exact and heuristic methods. Under the maximum time limit of 3,600s, Gurobi fails to provide an optimal solution even for the small-scale CLRP20 instances, and its solution quality remains inferior to that of our DRLHQ. The heuristic methods require significantly more time than our DRLHQ algorithm while yielding less competitive results. As to DRL-based methods, the increase in computational time is relatively small, while solution quality remains consistently high. Among all DRL-based methods, our DRLHQ exhibits the best solution quality across all cases, with no obvious inferiority in computation efficiency. For the CLRP100 instances, our DRLHQ delivers high-quality solutions in around 1 second. In contrast, exact and heuristic methods require more time and yield inferior results.

\subsubsection{Detailed comparison with DRL-based methods}
POMO and MTA are originally designed for generic vehicle routing problems and cannot directly solve the CLRP. To highlight that our method can be easily extended to other DRL methods, we integrate our dynamic masking mechanism into these algorithms, denoted by the mark ($^*$). For reference, we also report results from the vanilla versions to emphasize the importance of our work. As the results show, AM-W, POMO, and MTA exhibit a gap of no less than 9.36\% even on the CLRP10. As the problem scale increases to 100, the algorithm's performance degrades significantly, with a gap reaching 27.76\%. However, the algorithm's performance improves significantly across all problem scales by applying the dynamic masking mechanism proposed by our work. For example, on CLRP100, the gap of POMO* drops significantly from 27.76\% to 1.24\%. 

This phenomenon can be attributed to two main reasons: (1) decomposing the problem into two independently solved subproblems reduces solution optimality and neglects the correlation between heterogeneous decisions; (2) the decomposed subproblems differ significantly in size and node distribution, placing high demands on the generalization ability of DRL-based methods. It is clear that our DRLHQ outperforms all other DRL-based methods. Moreover, the proposed dynamic masking mechanism is easily transferable and achieves strong performance when applied to other DRL-based methods.

% \usepackage{multirow}
\begin{table*}[!t]
\caption{Results on Benchmark Datasets.}
\resizebox{\textwidth}{!}{
\begin{threeparttable} %添加此处
\begin{tabular}{c|c|cc|cc|cc|cc|cc|cc}
\toprule
\multirow{2}{*}{Inst.} & \multirow{2}{*}{BKS} & \multicolumn{2}{c|}{ILS} & \multicolumn{2}{c|}{SAH} & \multicolumn{2}{c|}{POMO$^*$} & \multicolumn{2}{c|}{MTA$^*$} & \multicolumn{2}{c|}{DRLHQ} & \multicolumn{2}{c}{DRLHQ-SGBS} \\
                     &                      & Cost        & Gap/BKS   & Cost         & Gap/BKS  & Cost         & Gap/BKS   & Cost         & Gap/BKS  & Cost          & Gap/BKS   & Cost            & Gap/BKS      \\ \midrule
20-5-1a              & 54793                & \textbf{56034}       & \textbf{2.26\%}    & 57421        & 4.80\%   & 56734        & 3.54\%    & 59288        & 8.20\%   & 57972         & 5.80\%    & 56045           & 2.28\%       \\
20-5-1b              & 39104                & 41721       & 6.69\%    & 43119        & 10.27\%  & 46269        & 18.32\%   & 43593        & 11.48\%  & 41415         & 5.91\%    & \textbf{41177}           & \textbf{5.30\%}       \\
20-5-2a              & 48908                & 67638       & 38.30\%   & 54403        & 11.24\%  & 51003        & 4.28\%    & 51060        & 4.40\%   & 51211         & 4.71\%    & \textbf{49186}           & \textbf{0.57\%}       \\
20-5-2b              & 37542                & \textbf{39624}       & \textbf{5.55\%}    & 42251        & 12.54\%  & 40823        & 8.74\%    & 40125        & 6.88\%   & 39751         & 5.88\%    & 39751           & 5.88\%       \\ \midrule
50-5-1               & 90111                & 108754      & 20.69\%   & 103565       & 14.93\%  & 97438        & 8.13\%    & 97089        & 7.74\%   & 95302         & 5.76\%    & \textbf{93087}           & \textbf{3.30\%}       \\
50-5-1b              & 63242                & 73138       & 15.65\%   & 74256        & 17.42\%  & 68201        & 7.84\%    & 68679        & 8.60\%   & 64383         & 1.80\%    & \textbf{63882}           & \textbf{1.01\%}       \\
50-5-2               & 88298                & 110712      & 25.38\%   & 102953       & 16.60\%  & 96650        & 9.46\%    & 95743        & 8.43\%   & 90684         & 2.70\%    & \textbf{89280}           & \textbf{1.11\%}       \\
50-5-2b              & 67340                & 77117       & 14.52\%   & 78784        & 16.99\%  & 75070        & 11.48\%   & 72571        & 7.77\%   & 72595         & 7.80\%    & \textbf{69318}           & \textbf{2.94\%}       \\
50-5-2bis            & 84055                & \textbf{88423}       & \textbf{5.20\%}    & 92769        & 10.37\%  & 94082        & 11.93\%   & 95517        & 13.64\%  & 92485         & 10.03\%   & 90309           & 7.44\%       \\
50-5-2bbis           & 51822                & 60451       & 16.65\%   & 57111        & 10.21\%  & 56755        & 9.52\%    & 56162        & 8.37\%   & 57634         & 11.22\%   & \textbf{56220}           & \textbf{8.49\%}       \\
50-5-3               & 86203                & 94279       & 9.37\%    & 106744       & 23.83\%  & \textbf{92315}        & \textbf{7.09\%}    & 100622       & 16.73\%  & 100727        & 16.85\%   & 100171          & 16.20\%      \\
50-5-3b              & 61830                & 67690       & 9.48\%    & 67636        & 9.39\%   & 65960        & 6.68\%    & 66678        & 7.84\%   & 66344         & 7.30\%    & \textbf{64397}           & \textbf{4.15\%}       \\ \midrule
100-5-1              & 275993               & 334886      & 21.34\%   & \textbf{295301}       & \textbf{7.00\%}   & 330400       & 19.71\%   & 331224       & 20.01\%  & 326570        & 18.33\%   & 321112          & 16.35\%      \\
100-5-1b             & 214392               & 256120      & 19.46\%   & \textbf{236688}       & \textbf{10.40\%}  & 248944       & 16.12\%   & 252865       & 17.95\%  & 248658        & 15.98\%   & 240619          & 12.23\%      \\
100-5-2              & 194598               & 258213      & 32.69\%   & \textbf{212195}       & \textbf{9.04\%}   & 304910       & 56.69\%   & 311147       & 59.89\%  & 310718        & 59.67\%   & 270152          & 38.83\%      \\
100-5-2b             & 157173               & 231676      & 47.40\%   & \textbf{171172}       & \textbf{8.91\%}   & 235875       & 50.07\%   & 242674       & 54.40\%  & 238442        & 51.71\%   & 222447          & 41.53\%      \\
100-5-3              & 200246               & 290930      & 45.29\%   & 221270       & 10.50\%  & 206417       & 3.08\%    & 208680       & 4.21\%   & 204501        & 2.12\%    & \textbf{202876}          & \textbf{1.31\%}       \\
100-5-3b             & 152586               & 225728      & 47.93\%   & 166301       & 8.99\%   & 159391       & 4.46\%    & 161680       & 5.96\%   & 161343        & 5.74\%    & \textbf{157829}          & \textbf{3.44\%}       \\
100-10-1             & 290429               & 363717      & 25.23\%   & 338153       & 16.43\%  & 356306       & 22.68\%   & 354749       & 22.15\%  & 314618        & 8.33\%    & \textbf{307429}          & \textbf{5.85\%}       \\
100-10-1b            & 234641               & 314971      & 34.24\%   & 296229       & 26.25\%  & 301654       & 28.56\%   & 293485       & 25.08\%  & 250895        & 6.93\%    & \textbf{245115}          & \textbf{4.46\%}       \\
100-10-2             & 244265               & 282717      & 15.74\%   & 270032       & 10.55\%  & 261057       & 6.87\%    & 270071       & 10.56\%  & 265173        & 8.56\%    & \textbf{250453}          & \textbf{2.53\%}       \\
100-10-2b            & 203988               & 219072      & 7.39\%    & 232939       & 14.19\%  & 216807       & 6.28\%    & 226678       & 11.12\%  & 220470        & 8.08\%    & \textbf{213348}          & \textbf{4.59\%}       \\
100-10-3             & 253344               & 340980      & 34.59\%   & 303292       & 19.72\%  & 278148       & 9.79\%    & 274682       & 8.42\%   & 271826        & 7.30\%    & \textbf{262193}          & \textbf{3.49\%}       \\
100-10-3b            & 204597               & 222607      & 8.80\%    & 244689       & 19.60\%  & 222907       & 8.95\%    & 222432       & 8.72\%   & 219884        & 7.47\%    & \textbf{216211}          & \textbf{5.68\%}       \\ \midrule
200-10-1             & 479425               & 626676      & 30.71\%   & 559427       & 16.69\%  & 547473       & 14.19\%   & 588099       & 22.67\%  & 528312        & 10.20\%   & \textbf{508435}          & \textbf{6.05\%}       \\
200-10-1b            & 378773               & 453302      & 19.68\%   & 456252       & 20.46\%  & 430128       & 13.56\%   & 436076       & 15.13\%  & 429332        & 13.35\%   & \textbf{414691}          & \textbf{9.48\%}       \\
200-10-2             & 450468               & 570573      & 26.66\%   & 523278       & 16.16\%  & 634215       & 40.79\%   & 578317       & 28.38\%  & 521621        & 15.80\%   & \textbf{518602}          & \textbf{15.13\%}      \\
200-10-2b            & 374435               & 536042      & 43.16\%   & 432274       & 15.45\%  & 461432       & 23.23\%   & 453492       & 21.11\%  & 439732        & 17.44\%   & \textbf{430091}          & \textbf{14.86\%}      \\
200-10-3             & 472898               & 627433      & 32.68\%   & 532378       & 12.58\%  & 499397       & 5.60\%    & 513916       & 8.67\%   & 501719        & 6.09\%    & \textbf{497571}          & \textbf{5.22\%}       \\
200-10-3b            & 364178               & 505390      & 38.78\%   & 415645       & 14.13\%  & 383092       & 5.19\%    & 391410       & 7.48\%   & 388762        & 6.75\%    & \textbf{379182}          & \textbf{4.12\%}       \\ \midrule
Average              & 197323           & 251554    & 23.38\%   & 226284   & 13.85\%  & 230662   & 14.76\%   & 231960   & 15.40\%  & 222436    & 11.85\%   & \textbf{215706}      & \textbf{8.46\%} \\ \bottomrule
\end{tabular}
\begin{tablenotes} %添加此处
      \footnotesize
		\item $\dagger$ `BKS' indicates the best-known solution values. `Cost' denotes the solution values. `Gap/BKS' is calculated by (Cost - BKS)/BKS. The best-performing method with the lowest objective value is in \textbf{bold}.
		\item * The method marked with * means that the method is adapted by the dynamic masking mechanism which is proposed in this work.
     \end{tablenotes} %添加此处
\end{threeparttable} %添加此处
}
\label{bench_clrp}
\end{table*}

\begin{figure}[!t]
	\centering
	\includegraphics[width=\columnwidth]{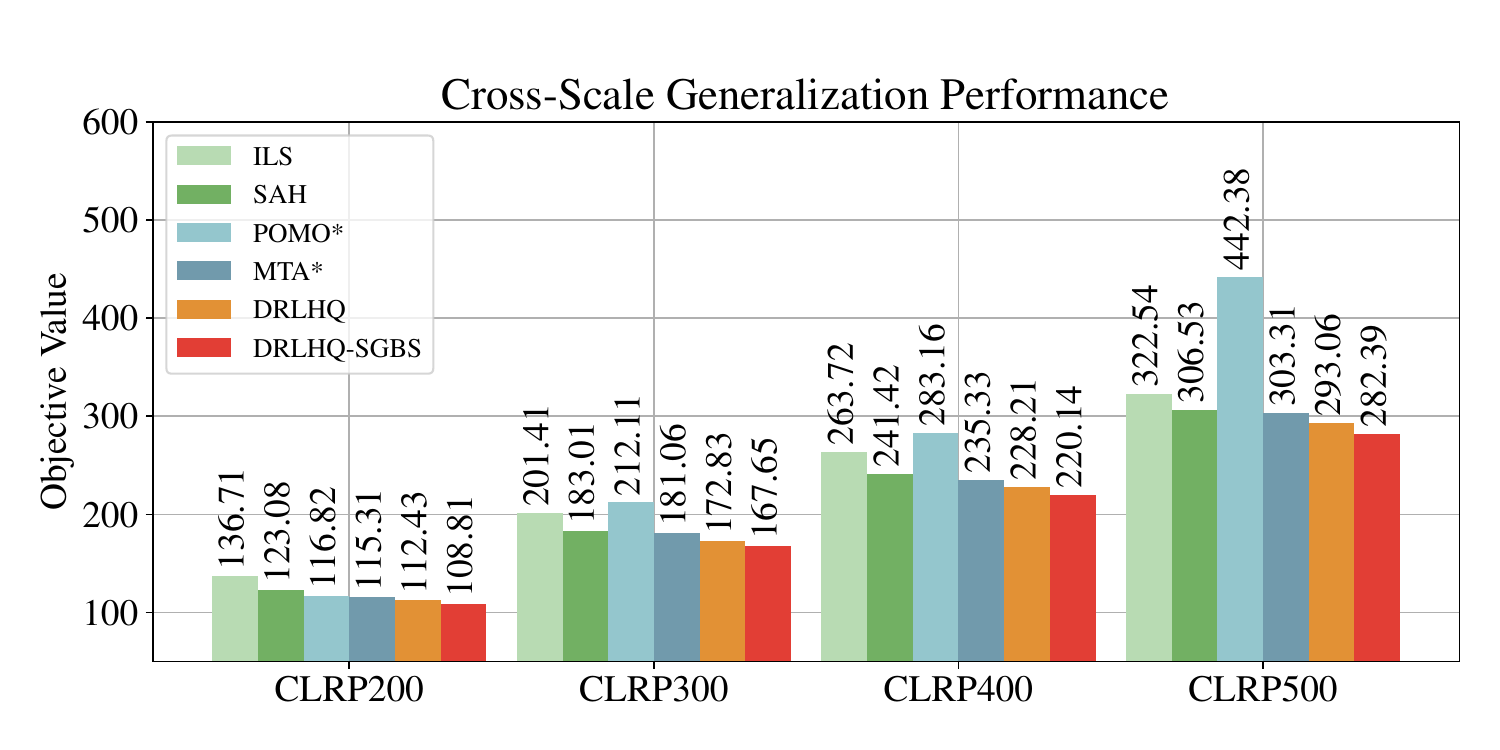}
	\caption{The generalization results on larger-scale CLRP instances, ranging from CLRP200 to CLRP500, show that our DRLHQ consistently outperforms all baseline methods across all problem scales.}
	\label{generalization}
\end{figure}

\subsection{Cross-Scale Generalization}
To evaluate the cross-scale generalization of our method, we further test DRLHQ on larger-scale instances. We consider four problem scales with the number of customers ranging from 200 to 500. For each scale, we randomly generate 100 instances. For the heuristic baselines (ILS and SAH), we use the same configurations as in the previous experiments to ensure a fair comparison. For DRL-based methods, including our DRLHQ and the adapted baselines POMO$^*$ and MTA$^*$, we directly apply the model pre-trained on CLRP100 to solve these larger-scale instances without any additional fine-tuning. Given the prohibitively high runtime of Gurobi on large-scale instances, we exclude it from this evaluation and only compare DRLHQ with ILS, SAH, POMO$^*$, and MTA$^*$.

As depicted in Fig. \ref{generalization}, DRL-based methods generally outperform heuristic baselines (except for POMO$^*$), and SAH consistently provides better solution quality than ILS. Among the DRL-based baselines, DRLHQ and MTA$^*$ achieve the best overall performance across all problem scales. However, the advantage of DRLHQ becomes more pronounced as the problem scale increases, leading to consistently superior solution quality on larger-scale instances.

Notably, despite being adapted with our dynamic masking mechanism, POMO$^*$ still shows a pronounced degradation as the scale increases. In contrast, DRLHQ remains much more stable across scales. This advantage can be attributed to our state-conditioned heterogeneous querying attention mechanism, which explicitly models and coordinates the coupled location and routing decisions. Since POMO$^*$/MTA$^*$ and DRLHQ share similar encoder-decoder backbones but differ in whether heterogeneous decision modeling is incorporated, these results further suggest that our proposed mechanism effectively improves cross-scale generalization on large-scale CLRP instances.

\subsection{Cross-Distribution Generalization}
\begin{figure}[!t]
  \centering
  \hspace{-.8em}
  \subfloat[Inst. 200-10-1\label{fig:a}]{
    \includegraphics[width=0.33\columnwidth]{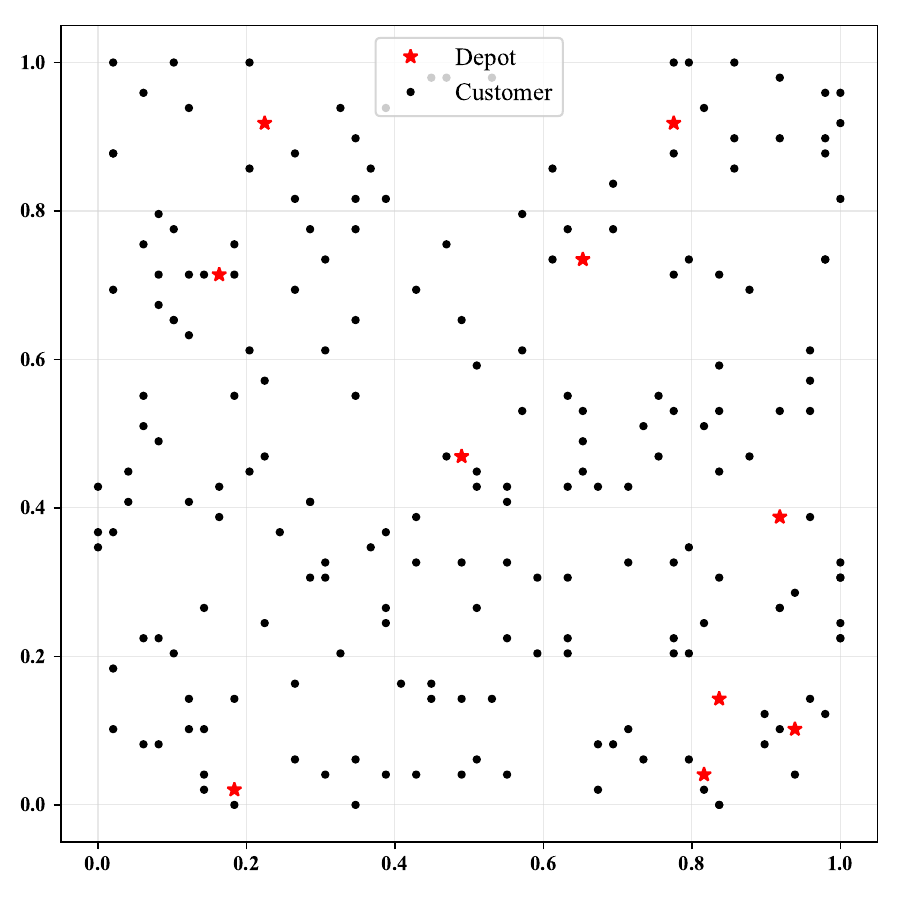}
  }
  \hspace{-1.2em}
  \subfloat[Inst. 200-10-2\label{fig:b}]{
    \includegraphics[width=0.33\columnwidth]{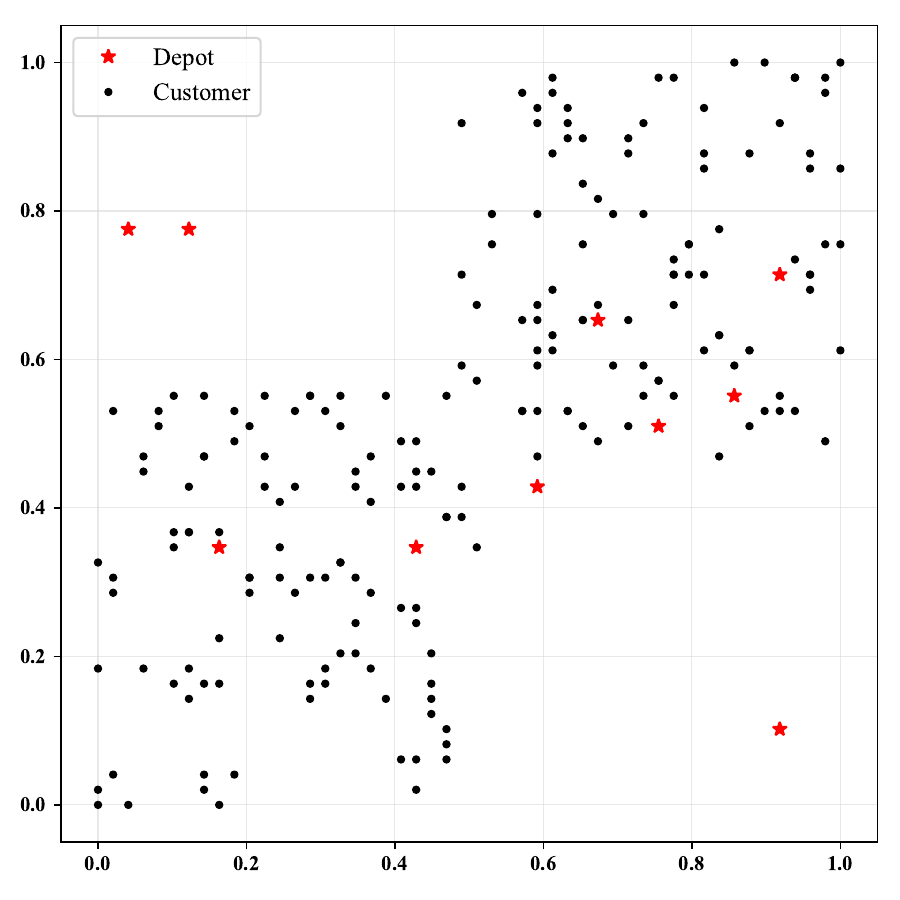}
  }
  \hspace{-1.2em}
  \subfloat[Inst. 200-10-3\label{fig:b}]{
    \includegraphics[width=0.33\columnwidth]{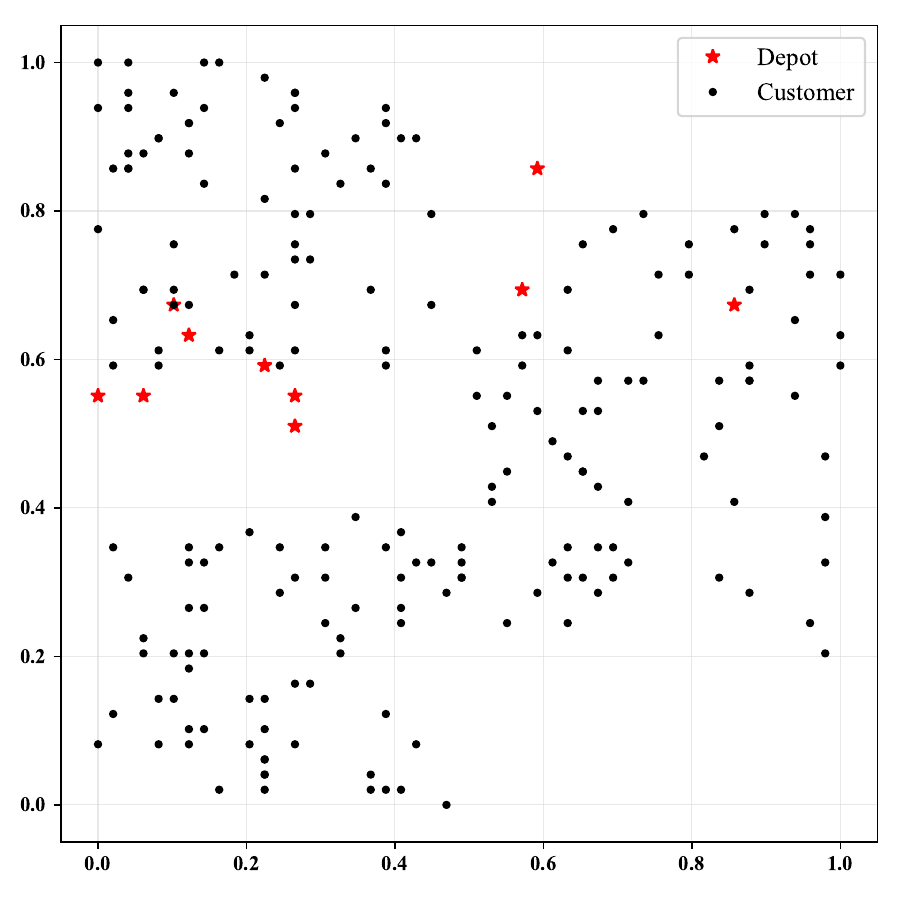}
  }
  \hspace{-.5em}
  \caption{The visualization for instances 200-10-1, 200-10-2, and 200-10-3, each containing 200 customers and 10 depots, which demonstrates significant variations in data distribution across different instances.}
  \label{fig:overall}
\end{figure}

We also evaluate the cross-distribution generalization performance of our method by applying it to public benchmark datasets. Specifically, we adopt the instances provided by Prins \textit{et al.} \cite{prins2004nouveaux}, which are publicly accessible and widely employed. It should be noted that these benchmark instances differ significantly from our synthetic datasets in several aspects, such as node distribution, customer demand, depot costs, and vehicle capacities. To better illustrate the differences in distribution, we visualize several representative instances, as shown in Fig. \ref{fig:overall}.

For benchmark datasets of varying scales, we evaluate each instance using the pre-trained model trained on synthetic datasets for all DRL-based methods. Specifically, for benchmark datasets with 20, 50, and 100 customer nodes, we directly use the corresponding models trained on CLRP20, CLRP50, and CLRP100, respectively. For larger-scale benchmark instances beyond our training scales, we apply the model trained on the CLRP100 synthetic dataset, without any additional fine-tuning. Table~\ref{bench_clrp} reports the performance comparison on benchmark CLRP instances of varying scales. 

Among heuristic methods, SAH generally achieves better solution quality than ILS, especially on larger-scale instances. Although ILS and SAH achieve reasonably good solution quality across all benchmark instances (with average Gap of 23.38\% and 13.85\%, respectively), as discussed before, they suffer from poor computational efficiency. Solving a single instance may take several hours, and their performance heavily relies on complex hand-crafted designs and parameter tuning, which limits their scalability and practical applicability.

For DRL-based baselines, POMO$^*$ and MTA$^*$ show competitive performance on small-scale instances, but their solution quality deteriorates significantly on larger benchmark instances. In contrast, DRLHQ consistently achieves lower gaps across all instance scales, demonstrating stronger robustness to complex node distributions and larger problem sizes. On average, DRLHQ reduces the gap to 11.85\%, outperforming all other baselines. Notably, when integrated with SGBS, DRLHQ-SGBS further improves solution quality across nearly all benchmark instances, achieving the best average gap of 8.46\%. 

Overall, these results demonstrate that DRLHQ can better capture the coupled location and routing decisions inherent in CLRP, leading to superior performance and stronger generalization compared with both heuristic and existing DRL-based methods on realistic benchmark datasets.

\subsection{Ablation Study}
As demonstrated by the previous experimental results, our DRLHQ achieves competitive performance across instances of various scales and datasets. To further validate the effectiveness of our method, we conduct an ablation study on the main components of DRLHQ. Specifically, DRLHQ comprises the encoder-decoder structure, the dynamic masking mechanism, and the state-conditioned heterogeneous querying attention mechanism. Since the encoder-decoder structure serves as the backbone of the algorithm and cannot be removed, we performed ablations by removing the dynamic masking mechanism and the state-conditioned heterogeneous querying mechanism individually to observe the corresponding changes in model performance. In the ablation study, we take the CLRP50 and CLRP100 as the testing cases. The results are shown in Table \ref{ablation_clrp}, where $Obj.$ indicates the objective values, while $Obj_v$, $Obj_d$, and $Obj_r$ correspond to the costs associated with vehicles, depots, and routing, respectively. 

%%%%%%%%%%%%%%%%%%%%

Built upon the basic encoder-decoder architecture, introducing the dynamic masking mechanism enables the model to explicitly capture the interdependency between location and routing decisions, leading to a substantial improvement in solution quality. Specifically, on CLRP50, the objective value is reduced from 37.2600 to 28.1288, while on CLRP100 it decreases from 70.2959 to 56.2868. Notably, the observed performance improvement mainly stems from a significant reduction in routing cost $Obj_r$, indicating that considering location and routing decisions independently may overlook the impact of routing structure on location choices, thereby severely degrading solution quality. This observation further highlights the effectiveness of our proposed mechanism in jointly modeling heterogeneous decision stages in an end-to-end manner.

Further incorporating the state-conditioned heterogeneous querying attention mechanism consistently brings additional performance gains. With heterogeneous querying attention enabled, DRLHQ achieves the best objective values on both scales, reducing the objective from 28.1288 to 27.7717 on CLRP50 and from 56.2868 to 55.9037 on CLRP100. Improvements are also observed across all objective components (i.e., $Obj_v$, $Obj_d$, and $Obj_r$), indicating that the proposed mechanism strengthens the model’s capability to coordinate heterogeneous decision stages and balance vehicle usage, depot selection, and routing costs in a unified framework.

In terms of computational efficiency, introducing the dynamic masking mechanism and the state-conditioned heterogeneous querying attention mechanism introduces only a moderate increase in inference time. Even for CLRP100, the full DRLHQ model produces high-quality solutions in approximately one second, confirming that the performance improvements are achieved with acceptable computational overhead.

Overall, the ablation results validate that both the dynamic masking mechanism and the state-conditioned heterogeneous querying attention are essential to DRLHQ. Their combined effect enables more effective modeling of heterogeneous decision stages in CLRP, leading to superior solution quality while maintaining practical inference efficiency.

% \usepackage{multirow}
\begin{table}[!t]
\caption{Ablation Study for Modules in DRLHQ}
\resizebox{\columnwidth}{!}{
\begin{threeparttable}
\begin{tabular}{c|ccc|ccccc}
\toprule
\multirow{2}{*}{Scale}&\multirow{2}{*}{En-Dec$^\dagger$} & \multirow{2}{*}{Masking$^\dagger$} & \multirow{2}{*}{HQ$^\dagger$} & \multicolumn{5}{c}{DRLHQ}                                                                  \\
                         &                                  &                          &                     & $Obj.$    & $Obj_v$ & $Obj_d$ & $Obj_r$ & Time (s)   \\ \midrule
\multirow{3}{*}{50}  & $\checkmark$                                &                          &                     & 37.2600 & 12.2380        & 6.1730       & 18.8490      & 0.2136 \\
                         & $\checkmark$                                & $\checkmark$                        &                     & 28.1288 & 11.3551        & 5.1420       & 11.6317      & 0.6186 \\
                         & $\checkmark$                                & $\checkmark$                        & $\checkmark$                   & \textbf{27.7717} & \textbf{11.2340}        & \textbf{5.2951}       & \textbf{11.2426}      & \textbf{0.6822} \\ \midrule
\multirow{3}{*}{100} & $\checkmark$                                &                          &                     & 70.2959 & 22.7200        & 16.1919      & 31.3840      & 0.5064 \\
                         & $\checkmark$                                & $\checkmark$                        &                     & 56.2868 & 21.9830        & 13.4195      & 20.8843      & 0.9374 \\
                         & $\checkmark$                                & $\checkmark$                        & $\checkmark$                   & \textbf{55.9037} & \textbf{21.7650}        & \textbf{13.4266}      & \textbf{20.7121}      & \textbf{1.0498}\\ \bottomrule
\end{tabular}
	\begin{tablenotes} %添加此处
      \footnotesize
		\item $\dagger$ `En-Dec' denotes the encoder–decoder architecture, `Masking' indicates the dynamic masking mechanism, and `HQ' represents the state-conditioned heterogeneous querying attention mechanism. The best-performing method with the lowest objective value is highlighted in \textbf{bold}.
     \end{tablenotes} %添加此处
\end{threeparttable} %添加此处
}
\label{ablation_clrp}
\end{table}

% \usepackage{multirow}

\section{Conclusion and Future Works}
\label{Conclusion}
\noindent In this paper, we propose an end-to-end DRL approach for solving the CLRP. Specifically, we reformulate the CLRP as an MDP that explicitly accommodates heterogeneous decision stages, yielding a general modeling framework that can be readily extended to other DRL-based methods. Based on the MDP transition rules, we introduce the dynamic masking mechanism to ensure solution feasibility throughout the construction process. To effectively capture the intricate interdependency between location and routing decisions, we adopt an encoder–decoder architecture and propose a state-conditioned heterogeneous querying attention mechanism in the decoder. Extensive experiments on both synthetic and benchmark datasets demonstrate that the proposed method consistently outperforms traditional heuristics and existing DRL-based baselines. Moreover, the results verify that our approach exhibits superior cross-scale and cross-distribution generalization capabilities. While applying DRL to CLRP remains relatively underexplored, our work provides empirical evidence of its effectiveness and highlights the promise of learning-based approaches for challenging COPs. For future work, we plan to investigate several directions: 1) extending the proposed framework to more complex routing problem variants \cite{liu2020two,wang2018hybrid,dorling2016vehicle}; 2) further enhancing cross-scale and cross-distribution generalization performance; and 3) incorporating uncertainty modeling to better address real-world application scenarios.

\bibliographystyle{IEEEtran}
\bibliography{references.bib}

\end{document}